\documentclass[a4paper,fleqn]{cas-dc}

\usepackage[authoryear,longnamesfirst]{natbib}

\def\tsc#1{\csdef{#1}{\textsc{\lowercase{#1}}\xspace}}
\tsc{WGM}
\tsc{QE}

\usepackage{pgfplots}
\usepackage{float}
\usepackage[below]{placeins}
\usepackage{tikz}
\usetikzlibrary{positioning, fit, arrows.meta, shapes, chains}
\usetikzlibrary{matrix, calc}
\usepackage{colortbl}

\usepackage[dvipsnames]{xcolor}
\usepackage{ifthen}
\definecolor{pinkrgb}{RGB}{180,60,150}
\definecolor{yellrgb}{RGB}{255,175,120}
\definecolor{bluergb}{RGB}{27,138,138}

\usepackage{caption}
\usepackage{subcaption}


\begin{document}
\let\WriteBookmarks\relax
\raggedbottom

\shorttitle{}    

\shortauthors{P. Pereira et al.}  

\title [mode = title]{Fuzzy Fingerprinting Encoder Pre-trained Language Models for Emotion Recognition in Conversations: Human Assessment and Validity Study}  



%

\author[1,2]{Patrícia Pereira}

\cormark[1]


\ead{patriciaspereira@tecnico.ulisboa.pt}


\credit{Conceptualization, Methodology, Investigation, Software, Formal analysis, Validation, Visualization, Writing – original draft, Writing – review \& editing, Funding acquisition}

\affiliation[1]{organization={Instituto Superior Técnico},
            addressline={Av. Rovisco Pais, 1}, 
            city={Lisbon},
            postcode={1049-001}, 
            country={Portugal}}

\author[3,2]{Helena Moniz}


\ead{helena.moniz@inesc-id.pt}


\credit{Conceptualization, Validation, Supervision, Writing – review \& editing}

\affiliation[2]{organization={INESC-ID},
            addressline={Rua Alves Redol, 9}, 
            city={Lisbon},
            postcode={1000-029}, 
            country={Portugal}}

\author[1,2]{Joao Paulo Carvalho}


\ead{joao.carvalho@inesc-id.pt}

\affiliation[3]{organization={Faculdade de Letras da Universidade de Lisboa},
            addressline={Alameda da Universidade}, 
            city={Lisbon},
            postcode={1600-214}, 
            country={Portugal}}

\credit{Conceptualization, Methodology, Validation, Supervision, Writing – review \& editing}

\cortext[1]{Corresponding author}



\begin{abstract}
In Emotion Recognition in Conversations (ERC), model decisions should align with nuanced human perception and ideally provide insights on the classification process. Standard encoder pre-trained language models (PLMs) are the state-of-the-art at these tasks but offer little insight into why a certain prediction is made. This is especially problematic in imbalanced datasets, where most utterances are labeled as neutral, making these models frequently misclassify minority emotions as the majority neutral class. To tackle this issue, we introduced a novel, interpretable approach to ERC by combining PLMs with Fuzzy Fingerprints (FFPs). FFP provide class-specific prototypes that reflect the characteristic class activation patterns in the PLM’s latent space. They are derived by ranking and fuzzifying the activations of the pooled conversational context-dependent embeddings across training instances for each emotion. At inference time, each input utterance is similarly fuzzy fingerprinted and matched to the emotion prototypes using a fuzzy similarity function based on the aggregation of the intersection of the fuzzy sets that define each FFP. Experimental results show that FFP integration reduces overclassification into the neutral class and human evaluation further supports the adequacy of FFP predictions. Our proposed method thus bridges the gap between deep neural inference and human perception, performing at state-of-the-art level while simultaneously offering valuable insights into the classification procedure.

\end{abstract}





\maketitle

\section{Introduction}\label{sec1}

Emotion Recognition in Conversations (ERC) is a challenging task in which the richness and ambiguity of human emotional expression require models that are simultaneously highly accurate and interpretable. ERC often operates in domains where human trust and psychological validity are key, such as empathetic dialogue agents, mental health monitoring, and therapeutic tools. Therefore, stakeholders must be able to understand the rationale behind a system’s classification of emotional states. However, most state-of-the-art ERC systems rely on deep neural architectures, particularly transformer-based models, that operate as so-called black boxes.
	
Furthermore, ERC datasets, such as DailyDialog \citep{li2017dailydialog} and MELD \citep{poria2018meld}, show severe label imbalance, with up to 83\% and 47\% of utterances labeled as \emph{Neutral} respectively, making state-of-the-art pre-trained language model (PLM) based classifiers misclassify minority inputs into this dominant class.
	
\begin{figure}[pos=h]
    \centering
        \includegraphics[width=\linewidth]{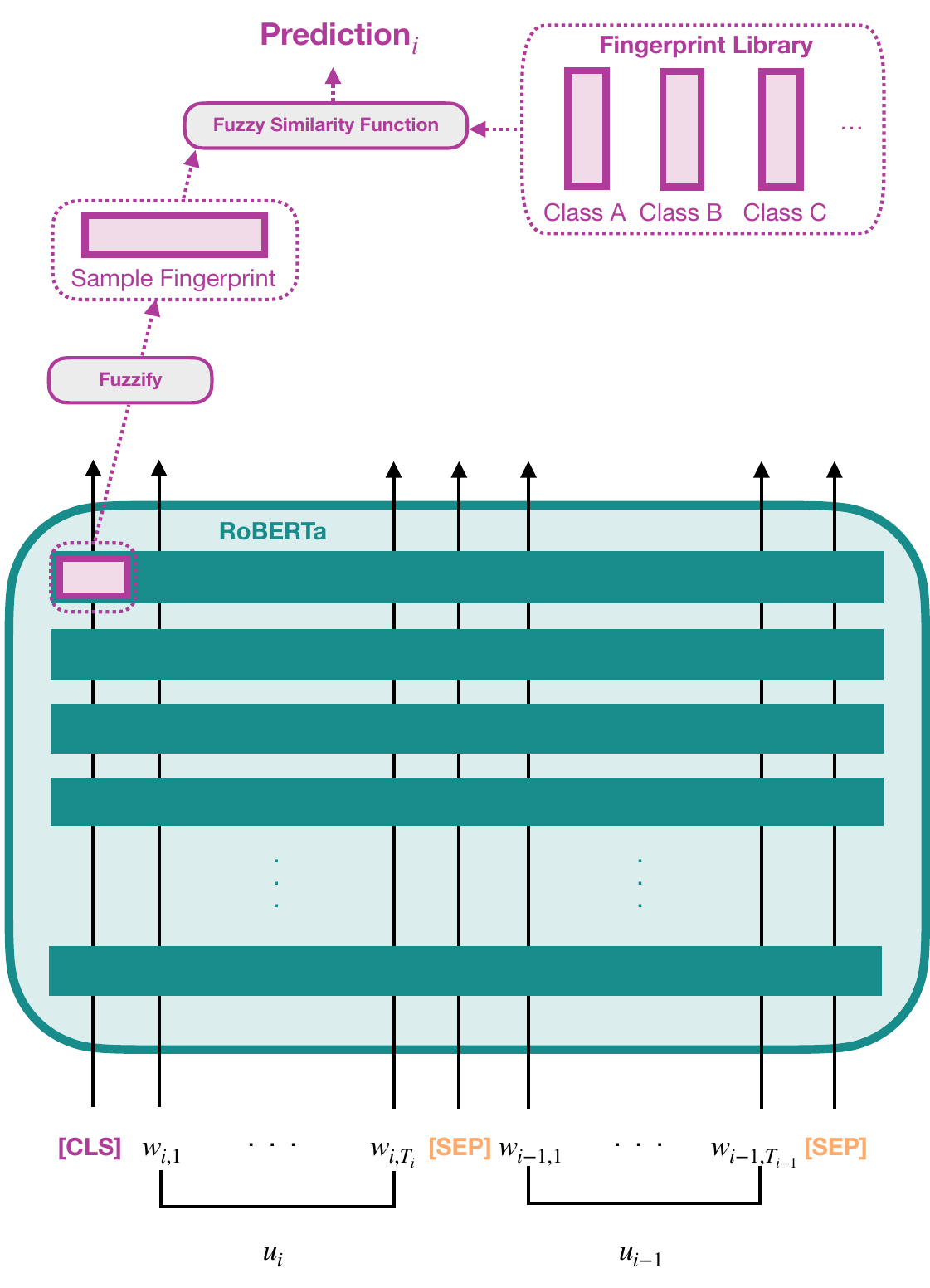}

    \caption{Model architecture. In this example input, an utterance, $u_{i}$, and its conversational context, $u_{i-1}$, are fed to the encoder PLM, of which the \texttt{[CLS]} token of the last layer goes through the Fuzzy Fingerprint module \citep{pereira2023fuzzy}.
    }
    \label{f1}
\end{figure}

To address both interpretability and performance in these settings, we have previously proposed integrating Fuzzy Fingerprints (FFPs) with PLMs \citep{pereira2023fuzzy}. A diagram of our approach is shown in Figure \ref{f1}. This hybrid architecture does more than just mitigate class imbalance. Rather than relying solely on softmax layers or attention weights, our method creates interpretable prototypes for each emotion, enabling instance-level explanations.
	
With FFPs, we can answer not just ``what did the model predict?'' but ``\textbf{why} did the model see this utterance as fear instead of sadness or neutral'', grounded in measurable similarity to a learned emotion fingerprint.
	
The classification insights of FFPs stem from their structure and transparency. Unlike softmax layers that entangle all weights into a single decision boundary, FFPs explicitly rank the importance of each latent feature for every class. This produces a prototype for each emotion that can be analyzed, visualized, and compared. The fuzzification process encodes degrees of relevance, differentiating features that are strongly associated with a particular emotion from those that are more ambiguous. The model does not just output a label, it allows us to state, for example, that ``this test utterance is most similar to the fingerprint for class A, due to high activation in the embedding cells X, Y, and Z.'' This reasoning mimics human categorization, where we judge new stimuli based on proximity to familiar prototypes.

To validate this observation, we conduct a human evaluation validity study in which we extract all instances where the FFP classifier produces a different label than the baseline model and perform an A/B test with human annotators concerning the labels generated by both models. Results indicate that the FFP classifier aligns more closely with human judgment. This suggests that beyond mitigating class imbalance, our approach enhances interpretability and produces classifications that better reflect human emotional perception. 

\section{Related Work}
\label{related}

\subsection{Emotion Recognition in Conversations}

ERC aims to assign an emotion label to each utterance in a dialogue. ERC is key for effective communication and is a crucial component in the development of responsive and socially aware systems. 

A key insight in ERC is that emotions are often shaped by context — they emerge and evolve across conversational turns. Therefore, leveraging the conversational context, i.e., the previous utterances, is crucial for accurate emotion prediction \citep{pereira2025deep}. Early neural approaches \citep{poria2017context} used LSTMs to encode the sequence of utterances, while DialogueRNN \citep{majumder2019dialoguernn} incorporated GRUs to track speaker-specific emotional states. These models explicitly considered dialogue structure, but their recurrent nature made it difficult to capture long-term dependencies.

The introduction of the Transformer \citep{vaswani2017attention} addressed this limitation through its self-attention mechanism and shorter path for gradient flow, enabling better modeling of the convesational context. Following this, encoder-based pre-trained Language Models (PLMs) such as BERT \citep{devlin2018bert}, RoBERTa \citep{liu2019roberta}, Longformer \citep{beltagy2020longformer}, and DeBERTa \citep{he2020deberta} became standard backbones for ERC tasks.

Recent ERC approaches have built on these PLMs not only for utterance encoding but also by incorporating external reasoning or knowledge. COSMIC \citep{ghosal2020cosmic} uses RoBERTa to encode utterances and supplements it with commonsense information extracted from a pretrained commonsense transformer model, along with five bi-directional GRUs. Psychological \citep{li2021past} also uses RoBERTa and a pretrained commonsense transformer model, while introducing a graph structure of utterances processed by a graph transformer. CoMPM \citep{lee2022compm} resorts to two RoBERTas in its proposed modules: a Context Embedding (CoM) module to process the conversational context, and a Pre-trained Memory (PM) module that encodes the previous utterances of the current speaker. SUNET \citep{song2023sunet} is based on a graph that has both speaker and utterance nodes, and different types of edges.

The recent surge in generative Large Language Models (LLMs) has brought new paradigms to classification tasks. These models, such as OpenAI’s GPT, Meta’s LLaMA, Mistral, and DeepSeek, allow for instruction-following and reasoning over multiple dialogue turns. In this setting, emotion classification is framed as a generative task: the LLM is prompted with an utterance, and possibly other relevant input, and asked to generate an appropriate label. While this strategy benefits from the general capabilities of LLMs, it also introduces new challenges: output variability, prompt sensitivity, and lack of transparency regarding the underlying reasoning process.

In contrast to all these approaches, our method leverages the representational power of PLMs while aiming for interpretability. We encode contextual utterances using RoBERTa \citep{pereira2023context}, and instead of using complex classification modules, we classify utterances based on their similarity to interpretable, class-specific Fuzzy Fingerprints derived from the PLM’s \texttt{[CLS]} output. This approach provides insights on which parts of the embeddings are most important for each emotion class.

\subsection{Fuzzy Fingerprints}

Fuzzy Fingerprints were introduced as a method for identifying a specific individual from a group of suspects in applications such as  Mobile User Identification \citep{homem2011mobile}, Web User Identification  \citep{homem2011web} or Text Authorship Identification  \citep{homem2011authorship}. In such approaches, Fuzzy Fingerprints are constructed based on the frequency of the task features. For authorship identification, a set of texts associated with a particular class is used to create the class (author) fingerprint, in which the frequency of each word in each text is used to build the fingerprint for that author. The fingerprint for a given class is the result of fuzzification and selection of the top-$k$ features, in this case, words, based on membership values that take into account their frequencies. The collection of the fuzzy fingerprints from all classes forms the fingerprint library. Given a fingerprint library and a test instance, the instance fingerprint is generated using the same FFP creation process, and a fuzzy-based similarity function is used to find the class with the most similar fingerprint. 

Conceptually, a FFP is a ranked, fuzzified set of features defining a prototype for a given class. A FFP is a fuzzy set in the discrete universe of the features, i.e., a set which contains elements with a varying degree of membership. In the original approach, the universe of features is potentially infinite and different for each class and each instance, despite all using $k$-sized fingerprints. Here we use a different approach \citep{carvalho2026fuzzy} where all FFP share the same limited universe of features.

The prototypical nature of FFP, where each class is associated with a Fingerprint, enables class and instance-level explanations for classification decisions.

\section{PLM Fuzzy Fingerprinting for ERC}
\label{methods}
\subsection{Task Definition}

Given a conversation composed of a sequence of $u_{i}$ utterances with corresponding $emotion_{i}$ from a predefined set of emotions, the aim of ERC is to correctly assign an emotion to each utterance of the conversation. An utterance consists of a sequence of $T_{i}$ tokens, $w_{it}$ for $T = 1, 2, ..., T_{i}$, as depicted on the bottom of Figure \ref{f1}.

\subsection{Context-Dependent Embedding Utterance Representations}

The most common approach for ERC has been to produce context-independent representations of each utterance (using PLMs), and subsequently perform contextual modeling of the obtained representations with classification modules comprising gated and graph neural networks. 

In a previous work \citep{pereira2023context}, we proposed to produce context-dependent representations of each utterance that represent not only the utterance but also a given number of previous utterances from the conversation. This context-based approach allowed us to discard the need for complex classification modules: a single fully connected linear softmax layer appended to this variation of the PLM, is enough to achieve state-of-the-art level performance. This earned us first place in a shared-task \citep{pereira2024context}.

We start by providing the PLM with the sentence we want to classify, denoted as $u_{i}$ concatenated with its conversation context - a certain number of previous sentences in the conversation, $u_{i-1}$, $u_{i-2}$, ..., up to $u_{i-c}$.

More specifically, we feed $u_{i}$ into the model, with the \texttt{[CLS]} token before it and the \texttt{[SEP]} token after it, followed by the previous turns $u_{i-1}$ up to $u_{i-c}$ with \texttt{[SEP]} tokens between them, as detailed in the diagram on Figure  \ref{f1}.

The encoder PLM produces multiple layers of embeddings that that can be used to represent the utterance and, in our proposed method, also the preceding utterances it receives as input. Each layer is composed of several tokens corresponding to the number of tokenized words in the segment. Each token is a vector with dimensions corresponding to the PLM's hidden size.

While these embeddings are all good candidates to represent the utterance and its context, opting for all tokens across all layers can lead to an excessively memory-intensive classification layer, compromising model performance and, more importantly, interpretability. Therefore, we select the first embedding from the final layer L, specifically the \texttt{[CLS]} token commonly utilized for classification.

This embedding is then fed to a fully connected linear layer with softmax so that the complete model maximizes the probability of the correct labels. 

For our FFP approach, we use the fully connected layer to fine-tune the PLM, but replace it afterwards with a Fuzzy Fingerprint classification module, as detailed in the next section and represented in Figure \ref{f1}.

\subsection{PLM Fuzzy Fingerprinting}
	
We adapt the concept of feature frequencies to the PLM's latent representation. Specifically, we adapt feature frequency analysis to the PLM’s final hidden state of the \texttt{[CLS]} classification token, which is a real-valued vector with a dimension $d$ that is fixed for each PLM. Although the vector's elements do not have an explicit semantic meaning, we treat the activation intensity of such elements as a proxy for their relevance within the PLM's latent representation, and use them as features for the creation of the FFP. This allows us to extract meaningful, class-specific patterns from the opaque latent space of the PLM. 
  
For a given emotion, the fingerprint is created by ranking and fuzzifying the $d$ features of the \texttt{[CLS]} embedding that are added across all training instances of that class. Highly activated features that consistently appear across the instances are therefore assigned higher fuzzy membership scores.

\subsection{FFP Creation}
\label{ffp_creation}

The procedure for creating the FFP for a given emotion is broken-down as follows:

\begin{enumerate}  

  \item 
  \textbf{Extract representations of training utterances:} The fine-tuned context-based PLM is fed with all the training examples of the given emotion (one by one).

  \item  
  \textbf{Aggregate representations:} The $d$-sized vectors generated by the PLM's output for each training example are accumulated into a $d$-sized fingerprint vector of real values. These aggregated embeddings represent how the PLM internally encodes that emotion.  

  \item 
  \textbf{Rank elements:} The aggregated values of the fingerprint vector are replaced by their activation rank, i.e., the feature with the highest accumulated value is replaced by "1", the second most activated feature is replaced by "2", and so on. The least activated feature will obviously assume the value $d$. This step identifies and ranks which latent features are more relevant to each emotion.
  
  \item 
  \textbf{Top-$k$ elements:} As in the original approach, the FFP only uses the top-$k$ more relevant features for classification purposes (instead of the whole vector of  $d$  features). The remaining ($d$-$k$) features are assigned a value of 0. This introduces sparsity and allows each fingerprint to be more readable and interpretable. $k$ is an hyperparameter that is tuned on a validation set.

  \item 
  \textbf{Fuzzify elements to obtain the FFP:} Each element of the fingerprint with a value larger than 0 is assigned a fuzzy membership value based on its rank. The following fuzzifying function is used for the experiments in this work:  
  \begin{equation}\label{mu}
  \mu_{i}=1-a \times \frac{i-1}{k}, i >0,\forall a \in [0,1]
  \end{equation} 
  in which $i$ is the value of the feature rank, $k$, is the fingerprint size, and $a$ adjusts the slope of the function (typically 1). Other functions were tested, this is the function that provided the best score.
      
\end{enumerate}

As a result, each class will have its own FFP, a vector of $d$ elements, of which $k$ elements have an associated membership value $]0,1]$. Note that the set of the top-$k$ elements of each class is usually different. Therefore, each FFP is a fuzzy set in the discrete universe of the $d$ features (e.g., 768 or 1024 in case the used PLM is RoBERTa).

\subsection{FFP Inference}  

After obtaining the FFP for all possible emotions (the FFP Library), classification can be performed. Given a utterance and its context to be classified:
  
\begin{enumerate}

  \item \textbf{Create the FFP of test utterance:} Use the above described procedure. In this case the FFP is created based on a single pass of the utterance and its context through the PLM  (i.e. rank the activations of the embedding representation, select the top-$k$ elements and fuzzify the resulting vector). As a result, one also obtains a fuzzy set in the discrete universe of the $d$ features, where only the top-$k$ most active features have a membership degree larger than zero. 
  \item \textbf{Compute similarity to emotion FFP:} Check the similarity of the test utterance FFP against the FFP of each emotion. The process is based on a very simple and efficient fuzzy sets operation: intersect the fuzzy sets of the uterance FFP with the fuzzy set of the Emotion FFP, and aggregate the resulting fuzzy set to obtain a crisp value. The process uses the Fuzzy Fingerprint similarity function from Equation \ref{eq:FFFcomparison},
  \begin{equation}
    \mbox{sim}(\Phi_E, \Phi_u) = \sum_{v = 1}^{d} \frac{{\min(\Phi_{Ev},\Phi_{uv})}}{N},
    \label{eq:FFFcomparison}
  \end{equation}
  where $\Phi_{xv}$ is the element $v$ of FFP $x$.  $N$ is an optional constant used for normalization purposes.
   \item \textbf{Select the emotion} with the highest similarity.
\end{enumerate}
  
\section{Experimental Setup}
\label{setup}

\subsection{Datasets}

Two famous benchmark datasets are used. DailyDialog \citep{li2017dailydialog} is built from websites used to practice English dialogue in daily life. The Multimodal EmotionLines Dataset (MELD) \citep{poria2018meld} contains dialogues from the TV series
Friends. Both datasets are labeled with the six Ekman’s basic emotions \citep{ekman1999basic}, \textit{Anger}, \textit{Disgust}, \textit{Fear}, \textit{Happiness}, \textit{Sadness} and \textit{Surprise}, or \textit{Neutral}. Table \ref{t0} resumes the main statistics of the datasets and Table \ref{t1} presents their label distribution. The publicly available splits of both datasets are used. 

\begin{table}[pos=h]
    \caption{Statistics of the datasets}
    \centering
    \begin{tabular}{lcc}
        \hline
        Dataset&DD&MELD\\
        \hline 
        Num. dialogues &13,118&1,433\\
        Num. turns &102,879&13,708\\
        Avg. turns per dialogue &7.8&9.6\\
        Avg. words per turn&14.6&8.0\\
        \hline
    \end{tabular} 
    \label{t0}
\end{table}

\begin{table}[pos=h]
    \caption{Proportion of labels in the datasets (\%)}
    \centering
  
    \begin{tabular}{lcccccccc}
        \hline
        & \textbf{Neu}&\textbf{Ang} & \textbf{Disg} & \textbf{Fear} & \textbf{Hap} & \textbf{Sad} & \textbf{Sur} \\ 
        DD&83.2&1.0&0.3& 0.1& 12.5 &1.1& 1.8 \\
        MELD&47.0&11.7 &2.6&2.6&16.8&7.3&11.9\\
        \hline
    \end{tabular} 
    \label{t1}
\end{table}

Because DailyDialog has a higher proportion of \textit{Neutral} samples than MELD, since the first is a dataset of conversations used for practicing English and the latter is from the Friends TV series, resorting to these two datasets provides an opportunity to test different domains and label distributions.

\subsection{Training Details}
\label{training_det}
\begin{table}[pos=h]
    \centering
    \caption{Training Parameters}
    \begin{tabular}{ll}
        \hline
        Parameter  &Value\\   
        \hline
        Model   &RoBERTa-base/large\\
        Loss Function &Cross-entropy loss\\
        Optimizer & Adam\\
        Initial Encoder Learning Rate& 1e-5\\
        Initial Head Learning Rate & 5e-5\\
        Encoder layer-wise decay rate  &0.95 per epoch\\
        Frozen Encoder Epochs &1st\\
        Batch Size &4\\
        Gradient Clipping &1.0\\
        Optimization Metric &Macro-F1\\
        Max \# of Training Epochs &10\\
        Early Stopping \# Epochs&5 \\
        \hline
    \end{tabular} 
    \label{tp}
\end{table}

From Table \ref{t1} it can be observed the both datasets are imbalanced, not only for their dominant majority neutral class but also for the relative imbalance between minority classes. To promote consistent performance across all classes we use the macro-F1 score for model selection. 

On Table  \ref{tp} we present the training parameters for obtaining the context-dependent embedding utterance representations from RoBERTa, the best performing PLM, prior to applying our FFP approaches. We also tested BERT, DeBERTa and Longformer, all in both base and large versions from the Transformers library by Hugging Face \citep{wolf-etal-2020-transformers}.

With regards to the parameter $a$ (Equation \ref{mu}) we have observed experimentally that 0.8 is a suitable value.

\subsection{Evaluation}

Our reported results are an average of 5 runs corresponding to 5 distinct random seeds that are kept for a meaningful comparison of all experiments. This is motivated by the fact that results for the same experiment obtained with different random seeds can have high variability in the macro F1-score, comparable to the improvements that we report upon state-of-the-art models. This 5 run average is in line with several ERC works.

\subsection{Generative LLM Experiments}

We also test LLMs, such as LLama and Mistral, specifically \texttt{Llama-3.3-70B-Instruct}, \texttt{Llama-4-Scout-17B-16E-Instruct}, \texttt{Emollama-chat-13}, and \texttt{Mistral-Small-3.1-24B-Instruct-2503}. 

We prompt the LLMs to  ``Analyze the following sentence and determine the primary emotion expressed, from the following: \texttt{[emotion categories]}. Output just the emotion. Consider the provided context. Context:  \texttt{[context sentences]} Sentence: \texttt{[sentence]}''.

\section{Results and Analysis}

\subsection{Fuzzy Fingerprint Size - $k$}

We start by analysing the effect of the FFP size $k$ on the performance (5 runs average). 

From the graphs in Figure \ref{kf} we can see that, for $k$ larger than 150 and 100 respectively, the performance of our proposed model is comparable to using all the 768 or 1024 RoBERTa outputs (and within state-of-the-art performance level). It is therefore possible to conclude that it is not necessary to use all the RoBERTa outputs to obtain high performance.

\begin{figure}[pos=h]
    \centering
    \resizebox{0.9\columnwidth}{!}{%
        \begin{tikzpicture}
            \begin{axis}[
                title={\textbf{DailyDialog}},
                xlabel={Fingerprint Size, $k$},
                ylabel={F1},
                xmin=0, xmax=800,
                ymin=10, ymax=55,
                legend pos=north west,
                ymajorgrids=true,
                grid style=dashed,
                grid=both,
                legend pos=south east,
                ylabel near ticks, yticklabel pos=left,
                ]

                \addplot[color=teal, mark=.] table[x=k, y=F1, col sep=semicolon, row sep=crcr, /pgf/number format/read comma as period] 
                {
                    k;F1\\
                    1;11,67\\
                    5;17,22\\
                    10;20,04\\
                    25;27,21\\
                    50;33,78\\
                    100;47,52\\
                    150;51,17\\
                    200;51,34\\
                    300;51,89\\
                    400;51,60\\
                    600;51,83\\
                    700;51,58\\
                    768;51,83\\
                }; 
                \addlegendentry{CD-ERC+FFP (Ours) \citep{pereira2023fuzzy}}
                
                \addplot[mark=none, orange, dashed, samples=2, domain = 0:800] {53.15};
                \addlegendentry{CoMPM \citep{lee2022compm}}
                
                \addplot[mark=none, violet, dashed, samples=2, domain = 0:800] {51.23};
                \addlegendentry{CD-ERC (Ours) \citep{pereira2023context}}
                
                \addplot[mark=none, magenta, dashed, samples=2, domain = 0:800] {51.05};
                \addlegendentry{COSMIC \citep{ghosal2020cosmic}}

            \end{axis}
        \end{tikzpicture}
    }
    \resizebox{0.9\columnwidth}{!}{%
        \begin{tikzpicture}
            \begin{axis}[
                title={\textbf{MELD}},
                xlabel={Fingerprint Size, $k$},
                ylabel={F1},
                xmin=0, xmax=1100,
                ymin=10, ymax=50,
                legend pos=north west,
                ymajorgrids=true,
                grid style=dashed,
                grid=both,
                legend pos=south east,
                ylabel near ticks, yticklabel pos=left,
                ]

                \addplot[color=teal, mark=.] table[x=k, y=F1, col sep=semicolon, row sep=crcr, /pgf/number format/read comma as period] 
                {
                    k;F1\\
                    1;10,40\\
                    10;32,26\\
                    50;40,54\\
                    100;46,32\\
                    200;46,54\\
                    300;46,88\\
                    400;46,74\\
                    500;46,68\\
                    600;46,84\\
                    700;47,01\\
                    800;47,13\\
                    1024;46,81\\
                }; 
                \addlegendentry{CD-ERC+FFP (Ours)}

                \addplot[mark=none, violet, dashed, samples=2, domain = 0:1100] {46.94};
                \addlegendentry{CD-ERC (Ours)}

            \end{axis}
        \end{tikzpicture}
    }
    \caption{Variation of the F1-score with $k$}
    \label{kf}
\end{figure}

The peak of performance happens for $k$=300 in DailyDialog and 800 in MELD, yielding F1-scores of 51.89 and 46.94 respectively, as presented on Table \ref{tkfv}.

\begin{table}[pos=h]
    \centering
    \caption{Variation of the F1-score with $k$}
    \begin{tabular}{ccccccc}
        \multicolumn{7}{c}{\textbf{DailyDialog}} \\
        \hline
        \textbf{K}& 1&5&10&25&50&100\\
        \textbf{F1}& 11.67&17.22&20.04&27.21&33.78&47.52\\
        \hline
        \textbf{K}&150&200&\textbf{300}&400&600&768\\
        \textbf{F1}&51.17&51.34&\textbf{51.89}&51.60&51.83&51.39\\
        \hline
        \multicolumn{6}{l}{\textbf{F1 without the Fingerprints module:}} &51.23\\
        \hline
   
        \noalign{\vspace{5pt}}
        \multicolumn{7}{c}{\textbf{MELD}} \\   
        \hline
        \textbf{K}& 1&10&50&100&200&300\\
    
        \textbf{F1}& 10.40&32.36&40.54&46.32&46.54&46.88\\
        \hline
        \textbf{K}&400&500&600&700&\textbf{800}&1024\\
        \textbf{F1}&46.74&46.68&46.84&47.01&\textbf{47.13}&46.81\\
        \hline
        \multicolumn{6}{l}{\textbf{F1 without the Fingerprints module:}} &46.94\\
        \hline
    \end{tabular} 
    \label{tkfv}
\end{table}

\subsection{Performance on each Emotion Label}

We proceed to analyse the Precision, Recall, and F1-score on each individual emotion label with the best value for $k$ in each dataset in Table \ref{tc} (5 runs average). 

\begin{table*}[]
    \centering
    \caption{Extensive Results}
    \begin{tabular}{lcccccccccccc}
        \hline
        & \multicolumn{3}{c}{Macro}& \multicolumn{3}{c}{Neu}&\multicolumn{3}{c}{Ang} & \multicolumn{3}{c}{Disg} \\
        &P&R&F1&P&R&F1&P&R&F1&P&R&F1 \\  
        \hline
        DD Baseline &54.95&49.29&51.23&90.71&92.15&91.42&47.50&40.34&43.51&49.43&25.11&33.22\\
        DD FFP &52.54&51.97&51.89&91.03&91.57&91.30&45.93&40.85&43.18&38.14&29.36&32.89\\
       
        MELD Baseline&49.86&46.04&46.94&75.60&81.20&78.13&53.20&44.35&47.55&34.04&23.24&27.42\\
        MELD FFP &47.97&47.84&47.13&76.57&78.84&77.63&53.02&42.72&47.00&29.46&28.82&28.80\\
        \hline
        &\multicolumn{3}{c}{Fear} & \multicolumn{3}{c}{Hap} & \multicolumn{3}{c}{Sad} & \multicolumn{3}{c}{Sur} \\
        &P&R&F1&P&R&F1&P&R&F1&P&R&F1 \\  
        \hline
        DD Baseline &43.21&36.47&39.44&63.27&59.19&61.12&42.30&35.88&38.43&48.25&55.86&51.50\\ 
        DD FFP &41.24&43.53&42.12&62.83&60.08&61.38&40.30&39.42&39.53&48.32&58.97&52.89\\
       
        MELD Baseline&24.19&21.20&22.01&58.41&59.70&58.65&48.54&31.63&37.37&55.00&61.00&57.37\\
        MELD FFP&20.35&31.20&23.75&61.03&58.41&59.48&43.31&32.69&36.97&52.03&62.21&56.38\\
          
        \hline
        
    \end{tabular} 
    \label{tc}
\end{table*}

The overall macro-F1 improves in both datasets, along with an increase in macro Recall, at the expense of a decrease in macro Precision. This may happen because the baseline model favours the dominant \textit{Neutral} class, yielding high Precision, whereas the FFP classifier assigns minority emotion labels more frequently, trading Precision for Recall.

In fact, the FFP classifier improves the Recall of the minority classes \textit{Fear}, \textit{Sadness}, \textit{Surprise} and \textit{Disgust} across both datasets.

Concerning the majority \textit{Neutral} class, the FFP classifier increases its Precision and decreases its Recall, which reflects the reduction of false-positive \textit{Neutral} predictions.

These results thus suggest the claimed lower overclassification of minority emotions into \textit{Neutral}.

We also present the confusion matrices of our approach on Figure \ref{confusions}.

\def\myConfMatdb{{
        { 5823.8,    42.4,    7.4,     6.4,   340 ,    48 ,    52 },  
        {   57.6,   47.6,     3.8,     0.8,     1,     2.2,     5 },  
        {  28.8,     5.6,   11.8,     0,     0,    0.4,     0.4},  
        {     10.4,     0.4,     0,   6.2,     0,   0,     0 },  
        {   399.2,     1,     0.8,     0.2,  602.6,     1.8,    12.4},  
        {   62 ,  1.8,     0.2,     0.6,     0.2,  36.6,    0.6  },  
        {   39 ,     1.8,    0,    0.2,   10.2,     0,  64.8},  
}}

\def\classNames{{"Neu","Ang" , "Disg" , "Fear" , "Hap" ,"Sad" , "Sur"  }} 

\def\numClasses{7} 

\def\myScale{1.05} 

\def\myConfMatdf{{
			{ 5788.4, 44.8, 14.2, 8.8, 352, 57.2, 55.6  },  
			{ 53.6, 48.2, 6, 1, 1, 3, 5.2 },  
			{ 27, 5.6, 13.8, 0 ,0 ,0.2, 0.4},  
			{   8.6, 0.4, 0.6, 7.4, 0 ,0 ,0},  
			{  389 ,1.4 ,1.2 ,0.6 ,611.6 ,1.8 ,12.4},  
			{ 57.6 ,2, 0.8 ,0.6, 0.2, 40.2, 0.6  },  
			{  34.4 ,2.8, 0 ,0 ,10.4, 0 ,68.4 },  
}}

\def\myConfMatmb{{
        { 1019 , 44 , 11.40  , 15 , 86.8 , 34.8 , 44  },  
        { 82.4  , 153 , 9.6  , 5.2 , 33.80 , 17.20  , 43.80 },  
        { 21.6 , 16.4 , 15.8 , 0.2 , 3.4 , 4 , 6.6},  
        {  18.8 , 4.2 , 0.8 , 10.6 , 3.2 , 6 , 6.4},  
        { 94.2, 24.6 ,4.20, 2, 240, 7.6, 29.40 },  
        {  76.6 ,22.2 ,2.8 ,10.2, 16.2, 65.8 ,14.2},  
        {  39.4 ,29.8, 2, 1.4 ,32.2, 4.8 ,171.4 },  
}}

\def\myConfMatmf{{
        { 989.4, 45.2 ,19.6 ,21.6 ,77.6 ,45.6 ,56  },  
        { 75.8, 147.4, 12.8, 11.4, 27.8, 22, 47.8 },  
        {21, 11.4, 19.6, 1.4, 2.4, 4.4, 7.8},  
        { 16, 2.2, 0.8, 15.6, 3.4, 6.2, 5.8},  
        { 85.6, 24.4, 5.8, 6.8, 234.8, 9.4, 35.20 },  
        { 71.2, 21.4, 6.6, 13.2, 15.2, 68, 12.4},  
        {34.8 ,28.6 ,2.6 ,6.2 ,25.8 ,8.2 ,174.8 },  
}}

\def\myConfMatdbd{{
        { 5823.8 , 496.2  },  
        { 597 ,  821},  
}}

\def\myConfMatdfd{{
	{ 5788.4 , 532.6 },  
	{ 570.2 , 847.8},  
}}

\def\myConfMatmbd{{
        { 1019, 236  },  
        { 333, 1021},  
}}

\def\myConfMatmfd{{
        { 989.4  ,265.6  },  
        { 304.4, 1049.6},  
}}

\def\classNamesd{{"Neu","Emo"  }} 

\def\numClassesd{2} 

\begin{figure*}[pos=!tp]
\centering
\begin{subfigure}[t]{.5\textwidth}
    \vspace{0pt}
    \centering
    \begin{tikzpicture}[
        scale = \myScale,
    ]
        \tikzset{vertical label/.style={rotate=90,anchor=east}}
        \tikzset{diagonal label/.style={rotate=45,anchor=north east}}
    
        \foreach \y in {1,...,\numClasses} 
        {
            \node [anchor=east] at (0.4,-\y) {\pgfmathparse{\classNames[\y-1]}\pgfmathresult};
    
            \foreach \x in {1,...,\numClasses} 
            {
                \def\totSamples{0}
                \foreach \ll in {1,...,\numClasses}
                {
                    \pgfmathparse{\myConfMatdb[\x-1][\ll-1] + \totSamples}
                    \xdef\totSamples{\pgfmathresult}
                }
    
                \begin{scope}[shift={(\x,-\y)}]
                    \pgfmathparse{\myConfMatdb[\x-1][\y-1]}
                    \let\mVal\pgfmathresult
                    \pgfmathtruncatemacro{\r}{\mVal}
                    \pgfmathtruncatemacro{\p}{round((\mVal/\totSamples)*100)}
    
                    \coordinate (C) at (0,0);
                    \ifthenelse{\p<50}{\def\txtcol{black}}{\def\txtcol{white}}
    
                    \node[
                        draw,
                        text=\txtcol,
                        align=center,
                        fill=teal!\p,
                        minimum size=\myScale*10mm,
                        inner sep=0,
                    ] (C) {\r\\\p\%};
    
                    \ifthenelse{\y=\numClasses}{
                        \node [] at ($(C)-(0,0.75)$)
                        {\pgfmathparse{\classNames[\x-1]}\pgfmathresult};
                    }{}
                \end{scope}
            }
        }
    \end{tikzpicture}
    \caption{DailyDialog (CD-ERC)}
    \label{conf-base}
\end{subfigure}%
\begin{subfigure}[t]{.5\textwidth}
    \vspace{0pt}
    \centering
    \begin{tikzpicture}[
        scale = \myScale,
    ]
        \tikzset{vertical label/.style={rotate=90,anchor=east}}
        \tikzset{diagonal label/.style={rotate=45,anchor=north east}}
    
        \foreach \y in {1,...,\numClasses} 
        {
            \node [anchor=east] at (0.4,-\y) {\pgfmathparse{\classNames[\y-1]}\pgfmathresult};
    
            \foreach \x in {1,...,\numClasses} 
            {
                \def\totSamples{0}
                \foreach \ll in {1,...,\numClasses}
                {
                    \pgfmathparse{\myConfMatdf[\x-1][\ll-1] + \totSamples}
                    \xdef\totSamples{\pgfmathresult}
                }
    
                \begin{scope}[shift={(\x,-\y)}]
                    \pgfmathparse{\myConfMatdf[\x-1][\y-1]}
                    \let\mVal\pgfmathresult
                    \pgfmathtruncatemacro{\r}{\mVal}
                    \pgfmathtruncatemacro{\p}{round((\mVal/\totSamples)*100)}
    
                    \coordinate (C) at (0,0);
                    \ifthenelse{\p<50}{\def\txtcol{black}}{\def\txtcol{white}}
    
                    \node[
                        draw,
                        text=\txtcol,
                        align=center,
                        fill=teal!\p,
                        minimum size=\myScale*10mm,
                        inner sep=0,
                    ] (C) {\r\\\p\%};
    
                    \ifthenelse{\y=\numClasses}{
                        \node [] at ($(C)-(0,0.75)$)
                        {\pgfmathparse{\classNames[\x-1]}\pgfmathresult};
                    }{}
                \end{scope}
            }
        }
    \end{tikzpicture}
    \caption{DailyDialog (CD-ERC+FFP)}
    \label{conf-ffp}
\end{subfigure}

\vspace{1.5em}

\begin{subfigure}[t]{.5\textwidth}
    \vspace{0pt}
    \centering
    \begin{tikzpicture}[
        scale = \myScale,
    ]
        \tikzset{vertical label/.style={rotate=90,anchor=east}}
        \tikzset{diagonal label/.style={rotate=45,anchor=north east}}
    
        \foreach \y in {1,...,\numClasses} 
        {
            \node [anchor=east] at (0.4,-\y) {\pgfmathparse{\classNames[\y-1]}\pgfmathresult};
    
            \foreach \x in {1,...,\numClasses} 
            {
                \def\totSamples{0}
                \foreach \ll in {1,...,\numClasses}
                {
                    \pgfmathparse{\myConfMatmb[\x-1][\ll-1] + \totSamples}
                    \xdef\totSamples{\pgfmathresult}
                }
    
                \begin{scope}[shift={(\x,-\y)}]
                    \pgfmathparse{\myConfMatmb[\x-1][\y-1]}
                    \let\mVal\pgfmathresult
                    \pgfmathtruncatemacro{\r}{\mVal}
                    \pgfmathtruncatemacro{\p}{round((\mVal/\totSamples)*100)}
    
                    \coordinate (C) at (0,0);
                    \ifthenelse{\p<50}{\def\txtcol{black}}{\def\txtcol{white}}
    
                    \node[
                        draw,
                        text=\txtcol,
                        align=center,
                        fill=teal!\p,
                        minimum size=\myScale*10mm,
                        inner sep=0,
                    ] (C) {\r\\\p\%};
    
                    \ifthenelse{\y=\numClasses}{
                        \node [] at ($(C)-(0,0.75)$)
                        {\pgfmathparse{\classNames[\x-1]}\pgfmathresult};
                    }{}
                \end{scope}
            }
        }
    \end{tikzpicture}
    \caption{MELD (CD-ERC)}
    \label{meld-conf-base}
\end{subfigure}%
\begin{subfigure}[t]{.5\textwidth}
    \vspace{0pt}
    \centering
    \begin{tikzpicture}[
        scale = \myScale,
    ]
        \tikzset{vertical label/.style={rotate=90,anchor=east}}
        \tikzset{diagonal label/.style={rotate=45,anchor=north east}}
    
        \foreach \y in {1,...,\numClasses} 
        {
            \node [anchor=east] at (0.4,-\y) {\pgfmathparse{\classNames[\y-1]}\pgfmathresult};
    
            \foreach \x in {1,...,\numClasses} 
            {
                \def\totSamples{0}
                \foreach \ll in {1,...,\numClasses}
                {
                    \pgfmathparse{\myConfMatmf[\x-1][\ll-1] + \totSamples}
                    \xdef\totSamples{\pgfmathresult}
                }
    
                \begin{scope}[shift={(\x,-\y)}]
                    \pgfmathparse{\myConfMatmf[\x-1][\y-1]}
                    \let\mVal\pgfmathresult
                    \pgfmathtruncatemacro{\r}{\mVal}
                    \pgfmathtruncatemacro{\p}{round((\mVal/\totSamples)*100)}
    
                    \coordinate (C) at (0,0);
                    \ifthenelse{\p<50}{\def\txtcol{black}}{\def\txtcol{white}}
    
                    \node[
                        draw,
                        text=\txtcol,
                        align=center,
                        fill=teal!\p,
                        minimum size=\myScale*10mm,
                        inner sep=0,
                    ] (C) {\r\\\p\%};
    
                    \ifthenelse{\y=\numClasses}{
                        \node [] at ($(C)-(0,0.75)$)
                        {\pgfmathparse{\classNames[\x-1]}\pgfmathresult};
                    }{}
                \end{scope}
            }
        }
    \end{tikzpicture}
    \caption{MELD (CD-ERC+FFP)}
    \label{meld-conf-ffp}
\end{subfigure}

\vspace{1.5em}

\begin{subfigure}[t]{.24\textwidth}
    \vspace{0pt}
    \centering
    \begin{tikzpicture}[
        scale = \myScale,
    ]
        \tikzset{vertical label/.style={rotate=90,anchor=east}}
        \tikzset{diagonal label/.style={rotate=45,anchor=north east}}
    
        \foreach \y in {1,...,\numClassesd} 
        {
            \node [anchor=east] at (0.4,-\y) {\pgfmathparse{\classNamesd[\y-1]}\pgfmathresult};
    
            \foreach \x in {1,...,\numClassesd} 
            {
                \def\totSamples{0}
                \foreach \ll in {1,...,\numClassesd}
                {
                    \pgfmathparse{\myConfMatdbd[\x-1][\ll-1] + \totSamples}
                    \xdef\totSamples{\pgfmathresult}
                }
    
                \begin{scope}[shift={(\x,-\y)}]
                    \pgfmathparse{\myConfMatdbd[\x-1][\y-1]}
                    \let\mVal\pgfmathresult
                    \pgfmathtruncatemacro{\r}{\mVal}
                    \pgfmathtruncatemacro{\p}{round((\mVal/\totSamples)*100)}
    
                    \coordinate (C) at (0,0);
                    \ifthenelse{\p<50}{\def\txtcol{black}}{\def\txtcol{white}}
    
                    \node[
                        draw,
                        text=\txtcol,
                        align=center,
                        fill=teal!\p,
                        minimum size=\myScale*10mm,
                        inner sep=0,
                    ] (C) {\r\\\p\%};
    
                    \ifthenelse{\y=\numClassesd}{
                        \node [] at ($(C)-(0,0.75)$)
                        {\pgfmathparse{\classNamesd[\x-1]}\pgfmathresult};
                    }{}
                \end{scope}
            }
        }
    \end{tikzpicture}
    \caption{DailyDialog (CD-ERC)}
    \label{2-dd-conf-baseline}
\end{subfigure}\hfill%
\begin{subfigure}[t]{.24\textwidth}
    \vspace{0pt}
    \centering
    \begin{tikzpicture}[
        scale = \myScale,
    ]
        \tikzset{vertical label/.style={rotate=90,anchor=east}}
        \tikzset{diagonal label/.style={rotate=45,anchor=north east}}
    
        \foreach \y in {1,...,\numClassesd} 
        {
            \node [anchor=east] at (0.4,-\y) {\pgfmathparse{\classNamesd[\y-1]}\pgfmathresult};
    
            \foreach \x in {1,...,\numClassesd} 
            {
                \def\totSamples{0}
                \foreach \ll in {1,...,\numClassesd}
                {
                    \pgfmathparse{\myConfMatdfd[\x-1][\ll-1] + \totSamples}
                    \xdef\totSamples{\pgfmathresult}
                }
    
                \begin{scope}[shift={(\x,-\y)}]
                    \pgfmathparse{\myConfMatdfd[\x-1][\y-1]}
                    \let\mVal\pgfmathresult
                    \pgfmathtruncatemacro{\r}{\mVal}
                    \pgfmathtruncatemacro{\p}{round((\mVal/\totSamples)*100)}
    
                    \coordinate (C) at (0,0);
                    \ifthenelse{\p<50}{\def\txtcol{black}}{\def\txtcol{white}}
    
                    \node[
                        draw,
                        text=\txtcol,
                        align=center,
                        fill=teal!\p,
                        minimum size=\myScale*10mm,
                        inner sep=0,
                    ] (C) {\r\\\p\%};
    
                    \ifthenelse{\y=\numClassesd}{
                        \node [] at ($(C)-(0,0.75)$)
                        {\pgfmathparse{\classNamesd[\x-1]}\pgfmathresult};
                    }{}
                \end{scope}
            }
        }
    \end{tikzpicture}
    \caption{DailyDialog (CD-ERC+FFP)}
    \label{2-dd-conf-ffp}
\end{subfigure}\hfill%
\begin{subfigure}[t]{.24\textwidth}
    \vspace{0pt}
    \centering
    \begin{tikzpicture}[
        scale = \myScale,
    ]
        \tikzset{vertical label/.style={rotate=90,anchor=east}}
        \tikzset{diagonal label/.style={rotate=45,anchor=north east}}
    
        \foreach \y in {1,...,\numClassesd} 
        {
            \node [anchor=east] at (0.4,-\y) {\pgfmathparse{\classNamesd[\y-1]}\pgfmathresult};
    
            \foreach \x in {1,...,\numClassesd} 
            {
                \def\totSamples{0}
                \foreach \ll in {1,...,\numClassesd}
                {
                    \pgfmathparse{\myConfMatmbd[\x-1][\ll-1] + \totSamples}
                    \xdef\totSamples{\pgfmathresult}
                }
    
                \begin{scope}[shift={(\x,-\y)}]
                    \pgfmathparse{\myConfMatmbd[\x-1][\y-1]}
                    \let\mVal\pgfmathresult
                    \pgfmathtruncatemacro{\r}{\mVal}
                    \pgfmathtruncatemacro{\p}{round((\mVal/\totSamples)*100)}
    
                    \coordinate (C) at (0,0);
                    \ifthenelse{\p<50}{\def\txtcol{black}}{\def\txtcol{white}}
    
                    \node[
                        draw,
                        text=\txtcol,
                        align=center,
                        fill=teal!\p,
                        minimum size=\myScale*10mm,
                        inner sep=0,
                    ] (C) {\r\\\p\%};
    
                    \ifthenelse{\y=\numClassesd}{
                        \node [] at ($(C)-(0,0.75)$)
                        {\pgfmathparse{\classNamesd[\x-1]}\pgfmathresult};
                    }{}
                \end{scope}
            }
        }
    \end{tikzpicture}
    \caption{MELD (CD-ERC)}
    \label{2-meld-conf-baseline}
\end{subfigure}\hfill%
\begin{subfigure}[t]{.24\textwidth}
    \vspace{0pt}
    \centering
    \begin{tikzpicture}[
        scale = \myScale,
    ]
        \tikzset{vertical label/.style={rotate=90,anchor=east}}
        \tikzset{diagonal label/.style={rotate=45,anchor=north east}}
    
        \foreach \y in {1,...,\numClassesd} 
        {
            \node [anchor=east] at (0.4,-\y) {\pgfmathparse{\classNamesd[\y-1]}\pgfmathresult};
    
            \foreach \x in {1,...,\numClassesd} 
            {
                \def\totSamples{0}
                \foreach \ll in {1,...,\numClassesd}
                {
                    \pgfmathparse{\myConfMatmfd[\x-1][\ll-1] + \totSamples}
                    \xdef\totSamples{\pgfmathresult}
                }
    
                \begin{scope}[shift={(\x,-\y)}]
                    \pgfmathparse{\myConfMatmfd[\x-1][\y-1]}
                    \let\mVal\pgfmathresult
                    \pgfmathtruncatemacro{\r}{\mVal}
                    \pgfmathtruncatemacro{\p}{round((\mVal/\totSamples)*100)}
    
                    \coordinate (C) at (0,0);
                    \ifthenelse{\p<50}{\def\txtcol{black}}{\def\txtcol{white}}
    
                    \node[
                        draw,
                        text=\txtcol,
                        align=center,
                        fill=teal!\p,
                        minimum size=\myScale*10mm,
                        inner sep=0,
                    ] (C) {\r\\\p\%};
    
                    \ifthenelse{\y=\numClassesd}{
                        \node [] at ($(C)-(0,0.75)$)
                        {\pgfmathparse{\classNamesd[\x-1]}\pgfmathresult};
                    }{}
                \end{scope}
            }
        }
    \end{tikzpicture}
    \caption{MELD (CD-ERC+FFP)}
    \label{2-meld-conf-ffp}
\end{subfigure}
\caption{Confusion Matrices. The vertical and horizontal axis correspond to the output and target classes, respectively.}
\label{confusions}
\end{figure*}

Upon examining the class-wise confusion matrices, it can be observed that classification errors exhibit different patterns that are not completely captured by the aggregate evaluation metrics.
 
Concerning the DailyDialog dataset, a great proportion of misclassified emotional utterances are assigned to the \emph{Neutral} class, while confusion between distinct emotional labels is generally less pronounced. For example, \emph{Fear} is predominantly misclassified as \emph{Neutral}, with very few instances being assigned to other emotional categories. This suggests that, for this dataset, a considerable part of the classification challenge lies in identifying the presence of an emotion rather than distinguishing between specific emotional categories. This observation should be interpreted in light of the severe class imbalance, particularly the very low representation of \emph{Fear} and \emph{Disgust}.

Concerning the MELD dataset, although \emph{Neutral} remains a frequent destination for misclassified emotional utterances, confusion between emotional categories is also more evident. For example, \emph{Fear} is not only classified as Neutral but is also assigned to \emph{Sadness}, \emph{Surprise}, \emph{Happiness} and \emph{Anger}. Therefore, the predominance of \emph{Neutral}-related errors observed in DailyDialog should not be generalized to all ERC datasets.

Upon examining the binary confusion matrices, it can be observed that the introduction of FFP consistently reduces the proportion of actual emotional utterances classified as \emph{Neutral}, from 42\% to 40\% in DailyDialog and from 25\% to 22\% in MELD, while the proportion of actual \emph{Neutral} utterances classified as emotional rounds to 8\% (although the count increases) and increases from 19\% to 21\%, respectively. This suggests that FFP favours the identification of emotional expressions, reducing false-negative emotion predictions at the cost of a moderate increase in false-positive emotional predictions, which is consistent with the increase in macro Recall and the decrease in macro Precision presented in Table \ref{tc}. 

The FFP classifier thus appears to favour the recognition of minority emotional classes, even when this entails a higher number of \emph{Neutral} utterances being assigned an emotional label. Importantly, these results also raise the question of whether all classification errors should be considered equally important when evaluating ERC systems.
	
\subsection{Ablation Studies}

We perform ablation experiments to analyze the individual contributions of each FFP computation step. Results are presented on Table \ref{t4} (5 runs average).

Prototype experiments correspond to using the aggregated values of the training samples (step 2 of subsection \ref{ffp_creation}) for direct comparison with the representation of the test sample, by calculating the dot product of their normalized values.

\begin{table}[pos=!htbp]
    \centering
    \caption{Ablation Studies}
    \tabcolsep=0.09cm
    \begin{tabular}{lcc}
        \hline
        &DailyDialog&MELD\\
        Setup&\textbf{macro-F1\%}&\textbf{macro-F1\%}  \\
        \hline
        Baseline (CD-ERC) &51.23 &46.94\\  
        Prototype &51.01 &43.15\\
        Prototype top-$k$&51.33 &47.10\\
        Fuzzy &51.39&46.81\\
        Fuzzy top-$k$ (CD-ERC+FFP) & 51.89&47.13\\
        \hline
    \end{tabular} 
    \label{t4}
\end{table}

It can be observed that standard Prototype experiments alone do not surpass the CD-ERC baseline, but the Prototype top$-k$ configuration (zeroing the non top-$k$ elements) beats the baseline. 

Concerning the Fuzzy configuration (corresponding to all FFP computation steps except the top-$k$ selection), it can be observed that these beat the DailyDialog baseline but not the MELD baseline. They also beat the Prototype configuration, and in DailyDialog also the Prototype top-$k$.

Finally, the complete CD-ERC+FFP model beats all of the aforementioned ablated models.

\subsection{Comparison with state-of-the-art}
\label{state}

Finally, we compare our approach to other state-of-the-art works that also resort to PLMs. This allows for a fair comparison between approaches given that using PLMs brings great performance increases when compared to using other means of utterance feature extraction. 

\begin{table*}[pos=!tp]

    \centering
    \caption{Comparison with state-of-the-art works}

    \begin{tabular}{lcccccc}
        \hline
        &\multicolumn{2}{c}{DailyDialog}&\multicolumn{2}{c}{MELD}\\
        &macro-F1\% &micro-F1\% &macro-F1\% &weighted-F1\%\\
        \hline
        RoBERTa \citep{ghosal2020cosmic} & 48.20&55.16 &-&62.02 \\
        RoBERTa + DialogueRNN \citep{ghosal2020cosmic} &49.65&57.32&-&63.61\\
        COSMIC \citep{ghosal2020cosmic} &51.05&58.48&-&65.21 \\

        SUNET \citep{song2023sunet}&- &59.48&-&64.03\\
        
         CoMPM \citep{lee2022compm}&\textbf{53.15}&\textbf{60.34}&-&\textbf{66.52} \\
         \hline
         CD-ERC (Ours) \citep{pereira2023context} &51.23&56.27&46.94&62.63\\
       
        CD-ERC+FFP (Ours) \citep{pereira2023fuzzy} &51.89& 56.44&47.13&62.94\\
        \hline
    \end{tabular} 
    \label{t5}
\end{table*}

We compare our approach to COSMIC \citep{ghosal2020cosmic}, RoBERTa and RoBERTa DialogueRNN, implemented by the authors of COSMIC, CoMPM \citep{lee2022compm} and SUNET \citep{song2023sunet}, all models described in Section \ref{related}. Results are displayed in table \ref{t5} (5 runs average).

We do not compare our approach to approaches that require knowledge of future utterances in the conversation \citep{li2021past}, since these are not suitable for real time ERC.

SUNET \citep{song2023sunet} has a higher performance than ours. It also leverages RoBERTa but does more computationally heavy graph construction, neighborhood attention mechanisms, and recurrent speaker-state updates, while we perform linear operations on the \texttt{[CLS]} token.

The CoMPM \citep{lee2022compm} model also has a higher performance than ours, but it resorts to two RoBERTa models. Furthermore, in the CoMPM reported ablation studies, when using only one of those modules (resorting to just one RoBERTa as our model does), the performance drops to 51.17 with the CoM module alone and 49.08 with the PM module alone, which is lower than our model's.

Our model leverages a single RoBERTa fine-tuned using context, and a minimalist fuzzy fingerprint classification model that can provide insights into the classification procedure, as elaborated in Section \ref{val}.

It is worth noting that we optimized macro-F1 in during training, as elaborated in subsection \ref{training_det}, so performance in micro-F1 and weighted-F1 can be even more competitive if the models are optimized for these metrics. 

Overall, performance on MELD is better than on DailyDialog, possibly due to the lower MELD imbalance between Neutral and the minority classes and also because MELD actors may have more expressive lines in the TV series Friends than the DailyDialog English learners.

\subsection{Generative LLM Experiments}

To complement our afoermentioned comparisons, we present the results of our experiments with Generative LLMs on Table \ref{tgen}. 

\begin{table}[pos=!htbp]
    \centering
    \caption{Generative LLM Results}
    \tabcolsep=0.07cm
    \begin{tabular}{lcc}
        \hline
        &DailyDialog&MELD\\
        Setup&\textbf{macro-F1\%} &\textbf{macro-F1\%} \\
        \hline
        \texttt{Emollama-chat-13}&23.68&25.54 \\  
        \texttt{Mistral-Small-3.1-24B-Instruct-2503}&26.99&37.07\\
        \texttt{Llama-3.3-70B-Instruct}&33.16&43.62\\
        \texttt{Llama-4-Scout-17B-16E-Instruct}&33.93&44.56\\
     
        \hline
    \end{tabular} 
    \label{tgen}
\end{table}

It can be observed that performance improves with model size. However, even the bigger models are still behind the tailored models for ERC from Table \ref{t5}.

Interestingly, these models often reassigned neutral utterances to emotional categories, echoing the behavior of our FFP classifier. This is the reason why performance on DailyDialog drops significantly. Given that its high proportion of \textit{Neutral} samples are now classified as emotional by the LLMs, \textit{Neutral} Recall drops significantly, leading to a drop in \textit{Neutral} F1, and subsequently macro-F1. But while in FFP, a mild reassignment occurs occurs within a supervised, task-tuned representation space with top-K prototype similarity, generative LLMs without task-specific calibration can over-predict emotional labels.

\section{Validity Studies}
\label{val}
Interpretability in neural classification is often an afterthought, relegated to post-hoc methods that try to make sense of opaque architectures after classification. When classifying solely with PLMs or other neural architectures, it is very difficult to discern the importance of each training example towards classification of the test instance since during training the model is adapting its weights to the example/emotion pairs but one cannot accurately measure the influence of each example neither emotion in the training nor classification procedure.

Our FFP approach breaks this mold by embedding interpretability into the classification mechanism. Rather than learning hidden decision boundaries, we construct visible, interpretable fingerprints for each emotion class and match input utterances against them during the classification procedure. 

\subsection{FFP Visualization}

Each fingerprint presented in Table \ref{ffp_examples} ($k$=7) represents a consensus activation pattern — a fuzzy set of PLM output cells ranked by their relevance to a given emotion. By distilling thousands of training examples into a compact, ranked list of salient cells, the fingerprint acts as a human-readable template for classification, allowing us to determine which PLM output cells are more important for a given emotion.

\begin{table*}[pos=h]
\centering
\begin{minipage}{\textwidth}
        \caption{Fingerprints ordered by rank and classification examples ($k$=7)}
        \centering
        \tabcolsep=0.09cm
        \begin{tabular}{lccccccc}
        
        \multicolumn{8}{c}{\textbf{DailyDialog}} \\
        \hline
        
        $FFP_{Neu}=$&
                \{(\textcolor{pinkrgb}{217},1),&(\textcolor{pinkrgb}{644},0.89),&(\textcolor{pinkrgb}{541},0.77),&(\textcolor{pinkrgb}{718},0.66),&(\textcolor{pinkrgb}{401},0.54),&(\textcolor{pinkrgb}{330},0.43),&(\textcolor{pinkrgb}{426},0.31)\}\\
        
        $FFP_{Ang}=$&
                \{(\textcolor{yellrgb}{8},1),&(\textcolor{yellrgb}{679},0.89),&(204,0.77),&(\textcolor{yellrgb}{292},0.66),&(651,0.54),&(573,0.43),&(111,0.31)\}\\
        
        $FFP_{Dis}=$&
                \{(588,1),&(573,0.89),&(27,0.77),&(154,0.66),&(331,0.54),&(67,0.43),&(561,0.31)\}\\
             
        $FFP_{Fear}=$&
                \{(588,1),&(313,0.89),&(655,0.77),&(406,0.66),&(736,0.54),&(349,0.43),&(624,0.31)\}\\
                
        $FFP_{Hap}=$&
                \{(588,1),&(585,0.89),&(388,0.77),&(600,0.66),&(767,0.54),&(319,0.43),&(741,0.31)\}\\
        
        $FFP_{Sad}=$&
                \{(371,1),&(588,0.89),&(5,0.77),&(156,0.66),&(4,0.54),&(93,0.43),&(550,0.31)\}\\
         
        $FFP_{Sur}=$&
                \{(691,1),&(588,0.89),&(97,0.77),&(573,0.66),&(530,0.54),&(535,0.43),&(654,0.31)\}\\
                
        \hline
        \noalign{\vspace{5pt}}
        \multicolumn{8}{c}{\textbf{MELD}} \\
        \hline
        
        $FFP_{Neu}=$&
                \{(529,1),&(986,0.89),&(659,0.77),&(584,0.66),&(734,0.54),&(483,0.43),&(543,0.31)\}\\
        
        $FFP_{Ang}=$&
                \{(788,1),&(672,0.89),&(1013,0.77),&(809,0.66),&(265,0.54),&(758,0.43),&(21,0.31)\}\\
        
        $FFP_{Dis}=$&
                \{(13,1),&(838,0.89),&(160,0.77),&(854,0.66),&(862,0.54),&(440,0.43),&(250,0.31)\}\\
             
        $FFP_{Fear}=$&
                \{(483,1),&(925,0.89),&(831,0.77),&(843,0.66),&(204,0.54),&(584,0.43),&(986,0.31)\}\\
                
        $FFP_{Hap}=$&
                \{(378,1),&(554,0.89),&(\textcolor{bluergb}{804},0.77),&(929,0.66),&(928,0.54),&(768,0.43),&(\textcolor{bluergb}{709},0.31)\}\\
        
        $FFP_{Sad}=$&
                \{(483,1),&(348,0.89),&(466,0.77),&(408,0.66),&(790,0.54),&(925,0.43),&(978,0.31)\}\\
         
        $FFP_{Sur}=$&
                \{(734,1),&(889,0.89),&(359,0.77),&(516,0.66),&(537,0.54),&(304,0.43),&(557,0.31)\}\\
                
        \hline
        \noalign{\vspace{5pt}}
        \multicolumn{8}{c}{\textbf{Classification Examples}} \\
        \hline

        Text: & \multicolumn{7}{l}{You still have not given me those files I ’ Ve asked you for .}\\
        
        $FFP_{Sample}=$&
                \{(\textcolor{yellrgb}{8},1),&(\textcolor{yellrgb}{679},0.89),&(309,0.77),&(624,0.66),&(\textcolor{yellrgb}{292},0.54),&(76,0.43),&(134,0.31)\}\\
        
        Similarity:&$Neu=0$ &$Ang=\textbf{0.35}$& $Dis=0$& $Fear=0$& $Hap=0$ &$Sad=0$ &$Sur=0$\\

        \hline
        Text: & \multicolumn{7}{l}{Don't forget to give me the files I've asked you for}\\
        $FFP_{Sample}=$&
                \{(\textcolor{pinkrgb}{330},1),&(\textcolor{pinkrgb}{644},0.89),&(\textcolor{pinkrgb}{541},0.77),&(\textcolor{pinkrgb}{217},0.66),&(114,0.54),&(\textcolor{pinkrgb}{426},0.43),&(211,0.31)\}\\
        
        Similarity:&$Neu=\textbf{0.44}$ &$Ang=0$& $Dis=0$& $Fear=0$& $Hap=0$ &$Sad=0$ &$Sur=0$\\
                
        \hline
        Text: & \multicolumn{7}{l}{You can do it!}\\
        $FFP_{Sample}=$&
                \{(\textcolor{bluergb}{709},1),&(329,0.89),&(379,0.77),&(146,0.66),&(425,0.54),&(\textcolor{bluergb}{804},0.43),&(211,0.31)\}\\
        
        Similarity:&$Neu=0$ &$Ang=0$& $Dis=0$& $Fear=0$& $Hap=\textbf{0.11}$ &$Sad=0$ &$Sur=0$\\
                
        \hline
        \end{tabular}
        
        \label{ffp_examples}
    
\end{minipage}
\end{table*}

Concerning the DailyDialog fingerprint, the first noticeable aspect is how output 588 is present at the fingerprints of 5 out of the 7 classes. This can be potentially leveraged to improve ERC performance using a multistage classifier, since 588 is not present in the Neutral FFP, which is the majority class and where most misclassifications end up.  
        
The first two examples show an interesting case of an incorrect classification and how the FFP can explain such error. The FFP of the utterance ``You still have not given me those files I've asked you for'' indicates a strong similarity to the emotion \textit{Anger}, with many important features in common between the fingerprints. While the utterance is labelled as \textit{Neutral}, it could certainly be argued that it has a negative meaning, revealing either an annotation error or some hidden knowledge from the annotator that cannot be extracted from the utterance. A more polite and neutral way of expressing the same message would be, for example,``Don't forget to give me the files I've asked you for'' (the other example on the table). Here we see that the FFP clearly indicates a \textit{Neutral} emotion despite the sentence expressing the same message, but with a different tone. The third example shows that even with a small $k$, a correct emotion can be detected by the non-zero intersection of the utterance ``You can do it!'' with only two elements of the fingerprint.
 
\begin{figure*}[pos=h]
    \centering
    \begin{minipage}[b]{0.48\textwidth}
        \centering
        \includegraphics[width=\linewidth]{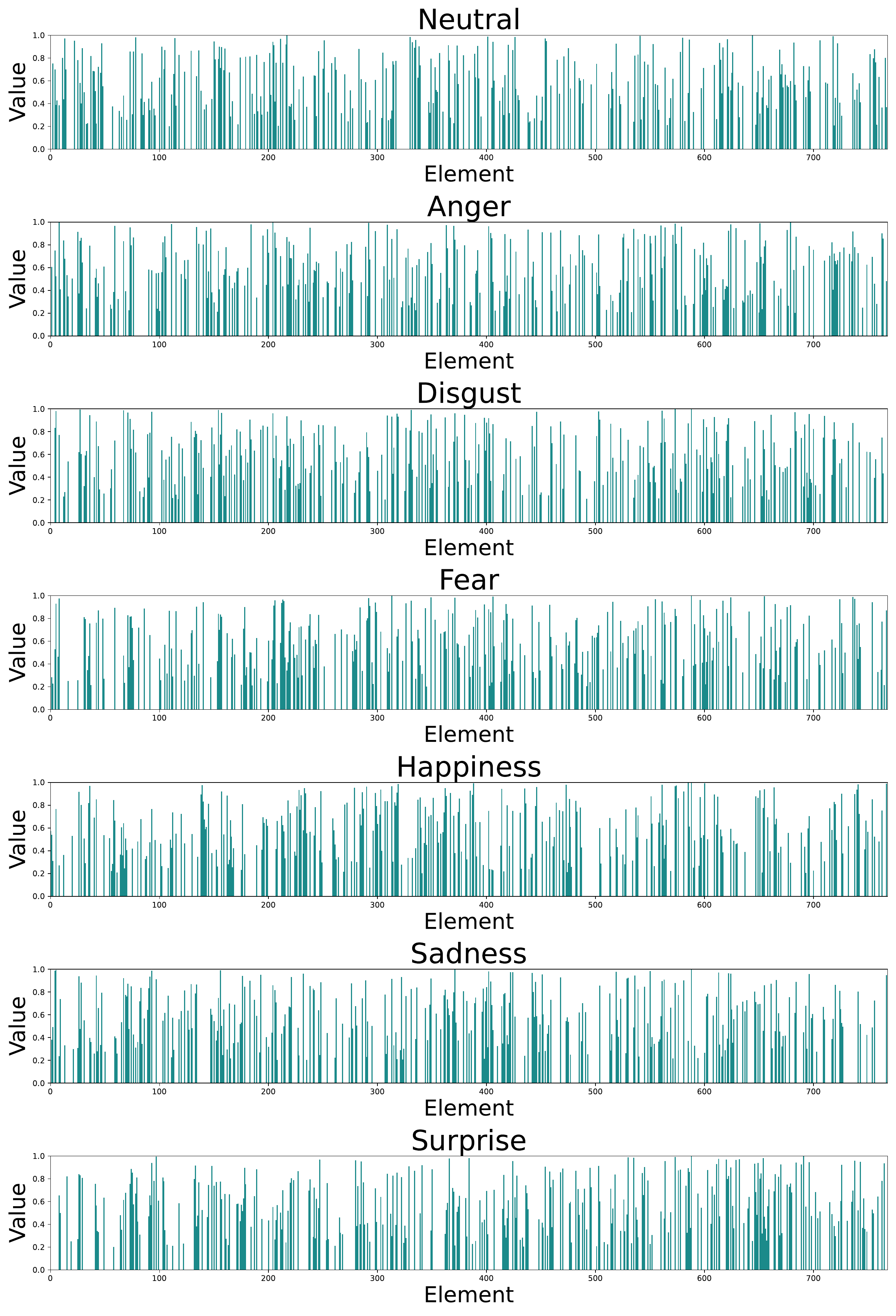}
        
    \end{minipage}
    \hfill
    \begin{minipage}[b]{0.48\textwidth}
        \centering
        \includegraphics[width=\linewidth]{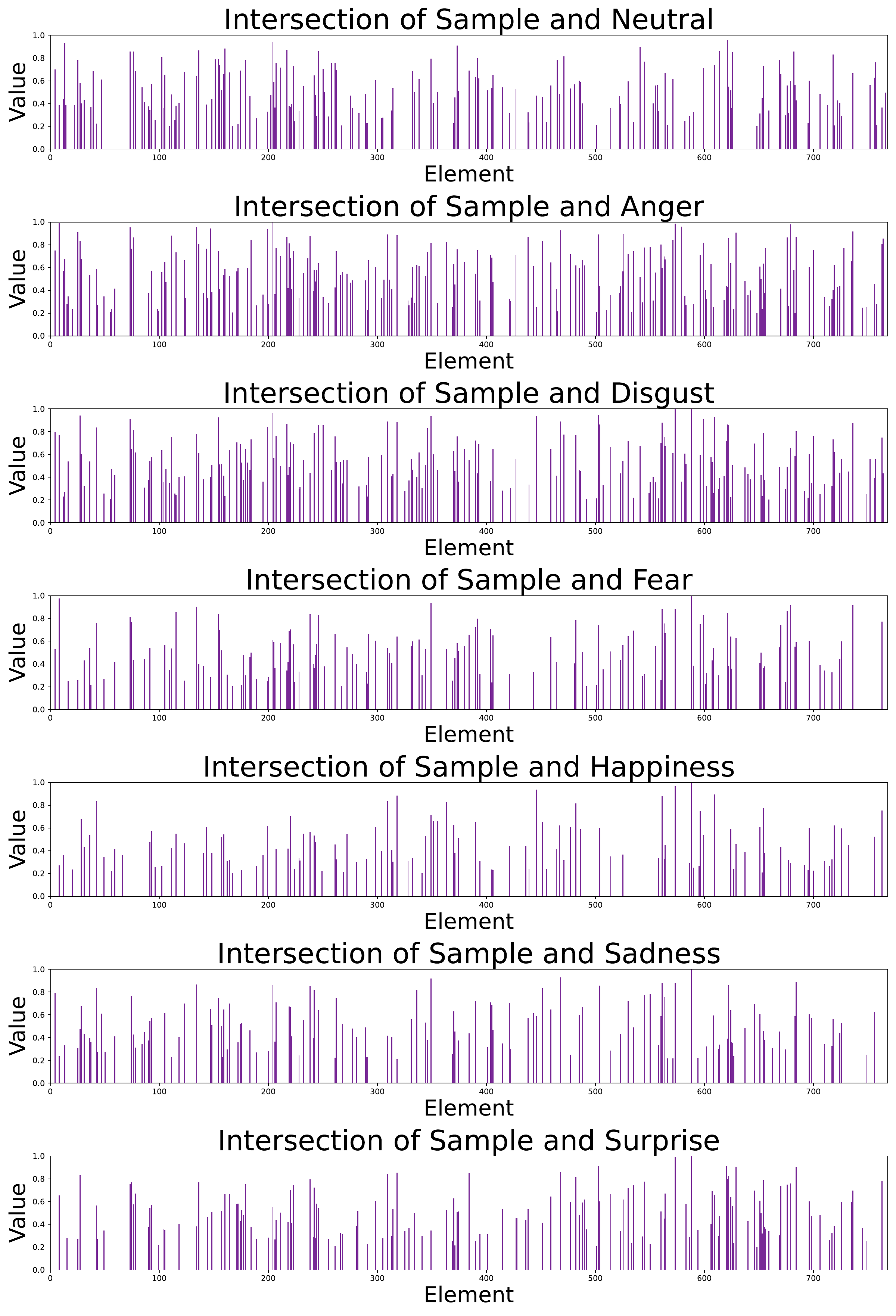}
           
    \end{minipage}
    
    \vspace{1.5em}
    
    \begin{minipage}[b]{0.48\textwidth}
        \centering
        \includegraphics[width=\linewidth]{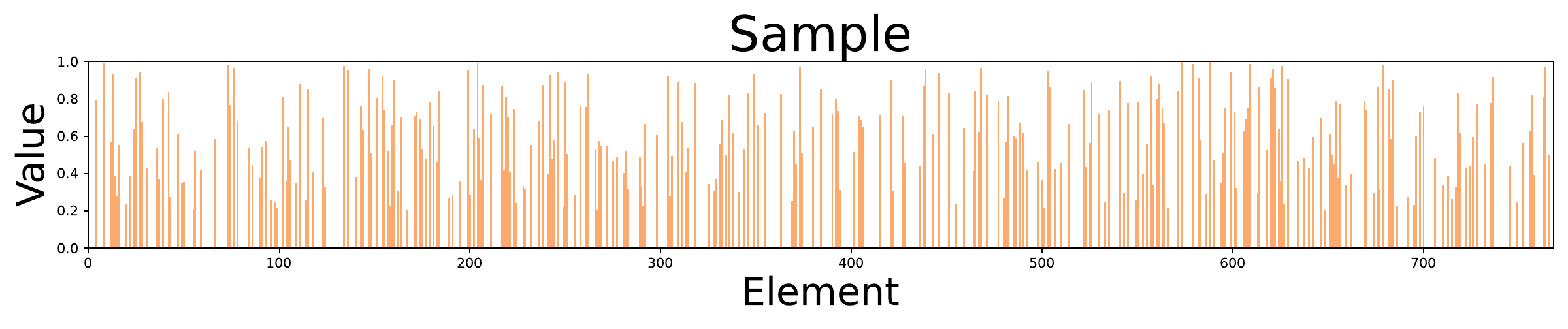}
        
    \end{minipage}
    \caption{DailyDialog FFP, sample, and intersection visualization}
    \label{show_dd}
\end{figure*}

Figure \ref{show_dd} ($k$=300) depicts for each DailyDialog emotion the unranked membership values of each element of the feature vector (RoBERTa's 768 cells), the fingerprint of the sample ``Whatever you say !'', and the intersection of this fingerprint with all the emotion class fingerprints.

It can be observed a higher density of filled element slots in the intersection with \textit{Anger}, as well as higher membership values. This is coherent with the higher class similarity score on Table \ref{class_ex}. There is a lower density and lower membership values in the intersection with other more distant emotions, as it is the case of \textit{Happiness}.
	  
\begin{figure*}[pos=h]
    \centering
    \begin{minipage}[b]{0.48\textwidth}
        \centering
        \includegraphics[width=\linewidth]{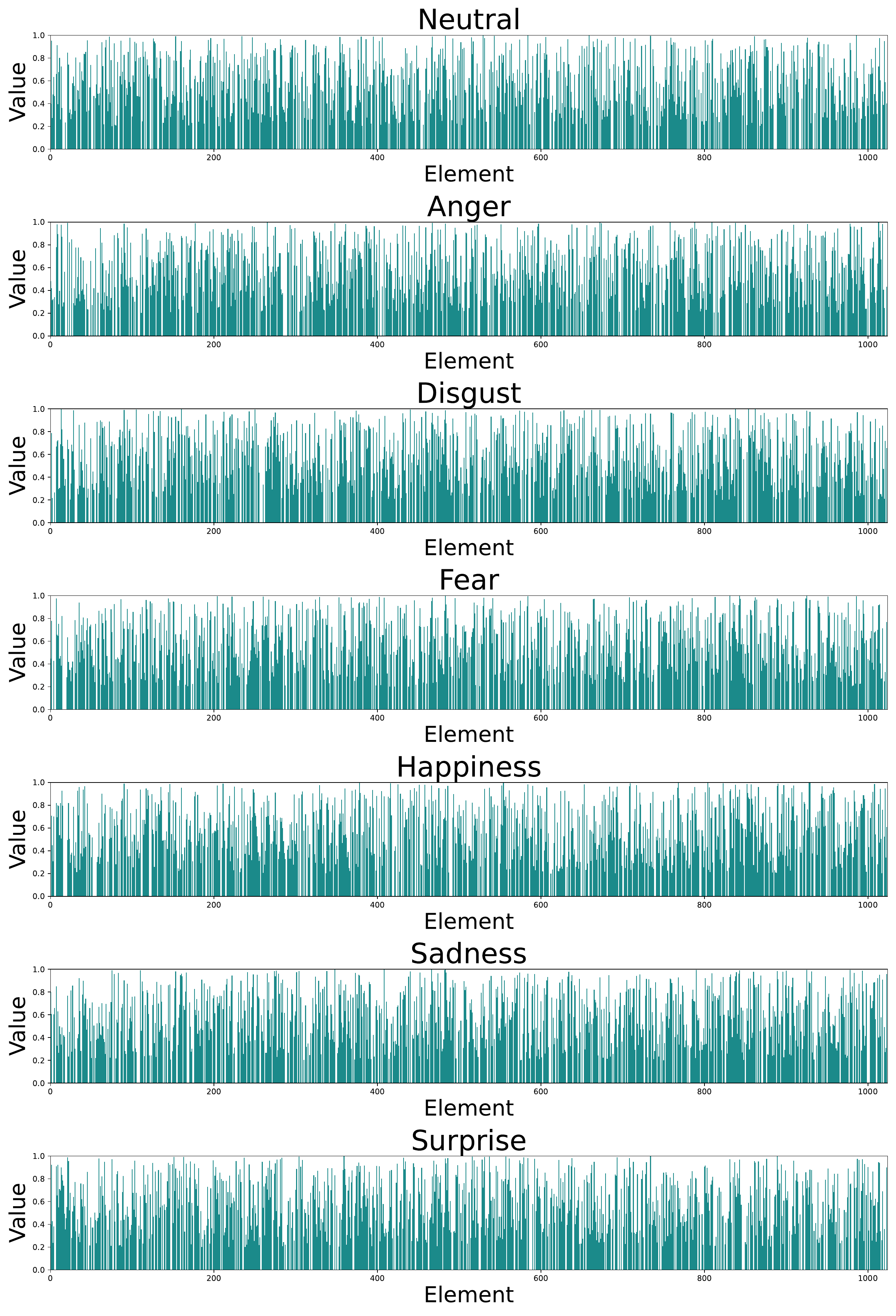}
      
    \end{minipage}
    \hfill
    \begin{minipage}[b]{0.48\textwidth}
        \centering
        \includegraphics[width=\linewidth]{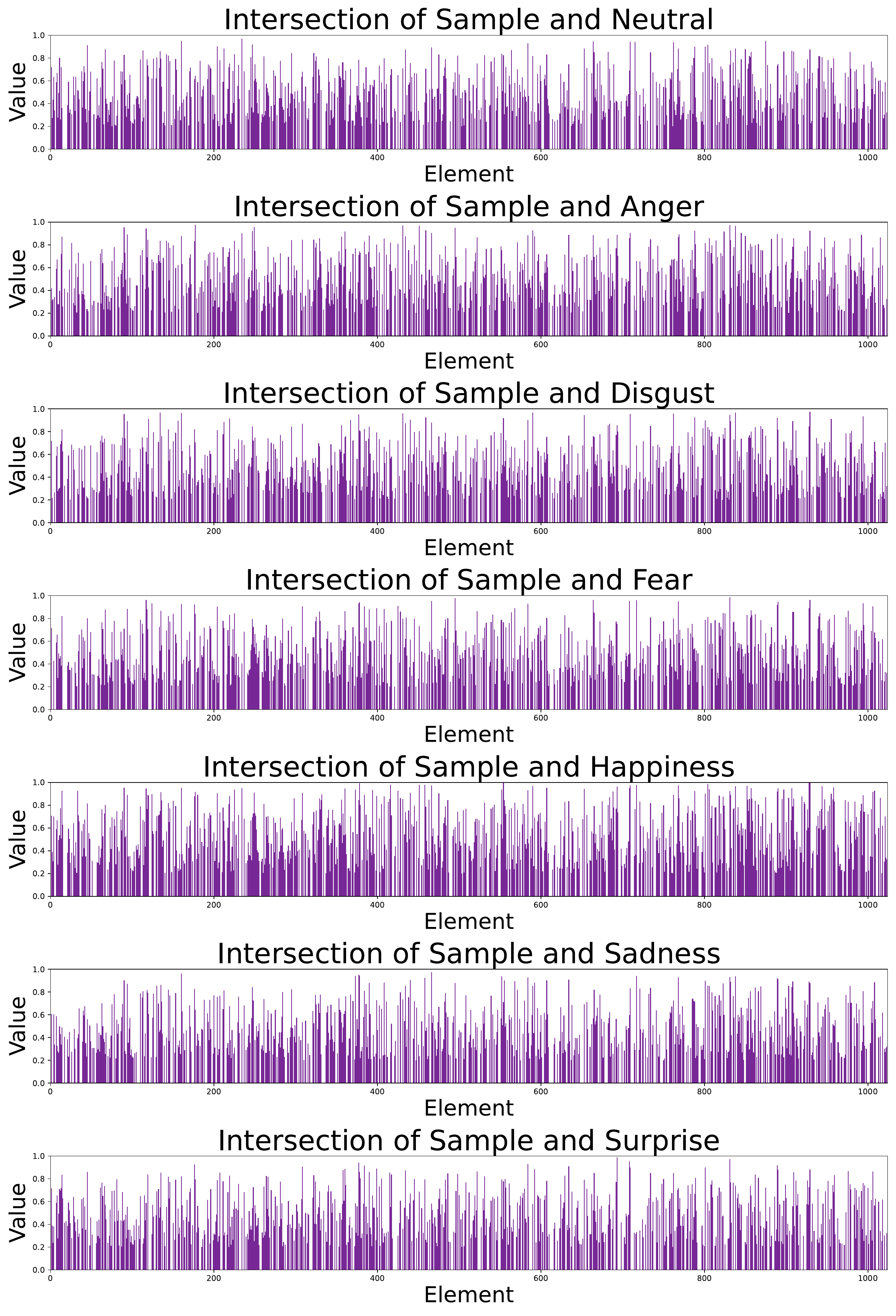}
   
    \end{minipage}
    
    \vspace{1.5em}
    
    \begin{minipage}[b]{0.48\textwidth}
        \centering
        \includegraphics[width=\linewidth]{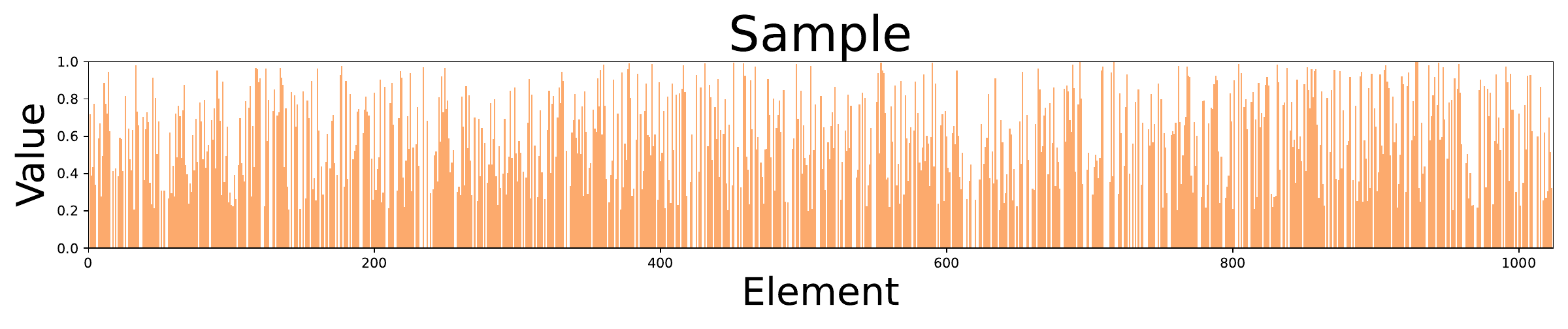}
      
    \end{minipage}
    \caption{MELD FFP, sample, and intersection visualization}
    \label{show_meld}
\end{figure*}

Figure \ref{show_meld} ($k$=800) depicts for each MELD emotion the unranked membership values of each element of the feature vector (RoBERTa's 1024 cells), the fingerprint of the sample ``Let's get that ball and really move, hey, hey, ho, ho.'', and the intersection of this fingerprint with all the emotion class fingerprints.

It can be observed a higher density of filled element slots in the intersection with \textit{Happiness}, as well as higher membership values. This is coherent with the higher class similarity score on Table \ref{class_ex}.

\subsection{Case Studies}
	\label{ffp_case}

We now present exemples of utterances and its class similarity scores (Equation \ref{eq:FFFcomparison}) on Table \ref{class_ex}.

In the first section of Table \ref{class_ex} we can observe examples from DailyDialog in which the FFP classifier yielded classifications that are in line with the dataset annotations, while the standard classifier yielded incorrect \textit{Neutral} classifications.

\begin{table*}[pos=h]
\centering
\begin{minipage}{\textwidth}
\caption{Classification examples}
    \centering
    \tabcolsep=0.09cm
    \begin{tabular}{lccccccc}
        
    \hline
    Text: & \multicolumn{7}{l}{Whatever you say!}\\

    Similarity:&$Neu=0.31$ &$Ang=\textbf{0.38}$& $Dis=0.35$& $Fear=0.25$& $Hap=0.18$ &$Sad=0.23$ &$Sur=0.26$\\
    
    \hline
    
    Text: & \multicolumn{7}{l}{I'm a little nervous.}\\
    
    Similarity:&$Neu=0.36$ &$Ang=0.24$& $Dis=0.27$&$Fear=\textbf{0.38}$& $Hap=0.18$ &$Sad=0.27$ &$Sur=0.22$\\

    \hline
    Text: & \multicolumn{7}{l}{This city is full of jerks}.\\

    Similarity:&$Neu=0.24$ &$Ang=0.34$& $Dis=\textbf{0.46}$ & $Fear=0.29$& $Hap=0.22$ &$Sad=0.26$ &$Sur=0.22$\\
    
    \hline
    \noalign{\vspace{5pt}}
    \hline
    
    Text: & \multicolumn{7}{l}{But it's true.}\\
    
    Similarity:&$Neu=0.23$ &$Ang=\textbf{0.39}$& $Dis=0.29$& $Fear=0.28$& $Hap=0.16$ &$Sad=0.30$ &$Sur=0.27$\\

    \hline
    
    Text: & \multicolumn{7}{l}{I wish the politicians would quit digging up dirt about each other's past.}\\
    
    Similarity:&$Neu=0.33$ &$Ang=0.32$& $Dis=\textbf{0.37}$& $Fear=0.25$& $Hap=0.15$ &$Sad=0.30$ &$Sur=0.24$\\

    \hline
    Text: & \multicolumn{7}{l}{No, this is not gonna happen! I... I've ruined everything...}\\

    Similarity:&$Neu=0.23$ &$Ang=0.25$& $Dis=0.27$& $Fear=0.27$& $Hap=0.25$ &$Sad=\textbf{0.37}$ &$Sur=0.19$\\

    \hline
    \noalign{\vspace{5pt}}
    \hline
    
    Text: & \multicolumn{7}{l}{Let's get that ball and really move, hey, hey, ho, ho.}\\
    
    Similarity:&$Neu=0.42$ &$Ang=0.43$& $Dis=0.42$& $Fear=0.42$& $Hap=\textbf{0.50}$ &$Sad=0.40$ &$Sur=0.40$\\

    \hline
    
    Text: & \multicolumn{7}{l}{Forgive me? You haven't been taking my calls in a week!}\\
    
    Similarity:&$Neu=0.39$ &$Ang=\textbf{0.55}$& $Dis=0.41$& $Fear=0.42$& $Hap=0.41$ &$Sad=0.42$ &$Sur=0.41$\\

    \hline
    \end{tabular}
    \label{class_ex}
\end{minipage}
\end{table*}

FFPs also help with auditing misclassifications. When the FFP classifier ``disagrees'' with the gold label, we can inspect the basis of its decision and potentially identify ambiguities in human annotation. In the second section of Table \ref{class_ex} we can observe examples from DailyDialog in which the FFP classifier yielded classifications that are not in line with the dataset annotations, and the model without the FFP yielded \emph{Neutral} classifications, according to the dataset. However, after reading the examples and confronting the class similarity scores, we observe that the FFP output is very reasonable and argue that the non-\emph{Neutral} assigned emotion would make sense. 

The third section of Table \ref{class_ex} presents examples from MELD dataset, in which the FFP classifier detected the correct emotional label.

\subsection{Human Evaluation Validity Study}
	
To assess which model aligns more closely with human emotional perception, we conducted an emotion annotation experiment with expert annotators on DailyDialog that was originally annotated by experts. 

We targeted all test instances where the FFP classifier and the baseline model without FFP produced different emotion labels for the same utterance, resulting in collecting 69 utterances. This design isolates cases of disagreement, which are the most informative for interpretability analysis. 
  
Each utterance was presented to three expert annotators in Computational Linguistics, all familiar with ERC tasks and discourse-level analysis, for an A/B Test.
  
\vspace{3 mm}
  
\textbf{Emotion A/B Test Guidelines:} You are presented with excerpts of dyadic (two-party) dialogues. For each excerpt, your task is to choose the emotion that best describes the speaker’s feeling in the final presented utterance (the bottom one).
Each final utterance is accompanied by up to 3 prior context turns to help you understand the conversation. These context turns may include the speaker of the final utterance (in blue) or the other participant (in pink). Each item will show 2 emotion options to choose from. These are randomized to avoid positional bias. For each item, choose an emotion option.
	
\vspace{3 mm}

\begin{figure}[pos=h]
   
    \centering
        \includegraphics[width=\linewidth]{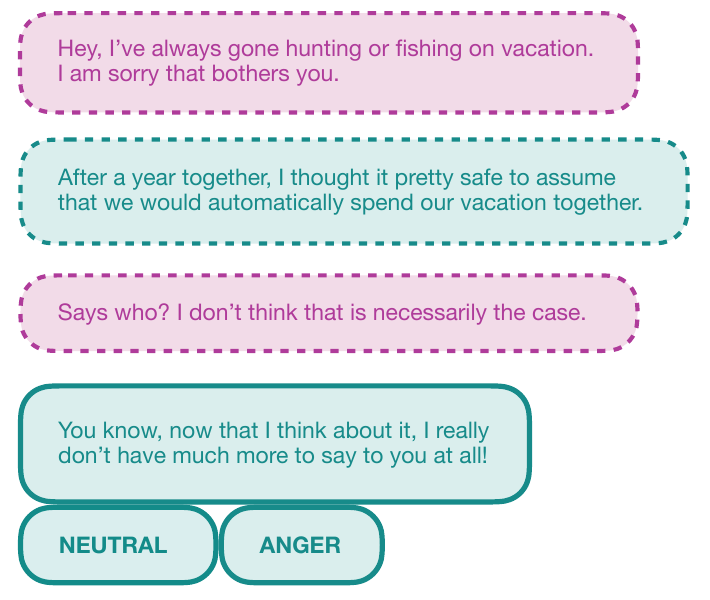}
    \caption{Annotation example}
    \label{ab2}
    
\end{figure}

We now present an annotation example on Figure \ref{ab2}, in which annotators favored \textit{Anger} over \textit{Neutral}, aligning with the label chosen by the FFP classifier. Their choice seems the most appropriate, since the speaker is clearly not satisfied with the fact that the other interlocutor is not joining for vacation, stating that ``I really don’t have much more to say to you at all!''.

Annotator preferences were towards the FFP classifier: the gold annotator preferences matched 59\% of the FFP classifier labels versus 41\% of the standard classifier labels. Their inter-annotator agreement, a Fleiss' Kappa of $0.47$, is very high considering that it is a typical value for emotion annotation, but in this case we isolated the most difficult and disagreement originating cases. 

These results indicate that FFP classifications align more often with human perception, particularly in emotionally nuanced or context-heavy utterances. Like expert annotators, the FFP classifier shows better insight into the characteristic patterns of each class, possibly due to its prototypical nature. This capability is challenging to achieve, especially with the high number of classes at stake.

This study supports the claim that FFPs provide not only quantitative improvements (as shown in performance metrics) but also qualitative alignment with human emotional perception. In tasks where subjective interpretation and contextual nuance are critical this insight is particularly valuable.

\section{On Application-Dependent Evaluation of ERC}

Motivated by the analysis of the confusion matrices, we present reflections for discussion on the performance analysis of ERC systems.

In conversational applications, the consequences of different classification errors may vary considerably. Misclassifying an emotional utterance as \emph{Neutral} may result in the system failing to acknowledge an emotional cue, leading to a less personalized or empathetic response. Conversely, assigning an incorrect emotional category may lead to a response that is inappropriate for the interlocutor's actual emotional state.  The relative importance of these errors is application-dependent. While failing to identify an emotional expression may be undesirable in general conversational agents, failing to detect \emph{Fear} or \emph{Sadness} may have more significant consequences in applications related to mental health support. 

So a model with higher Recall is not necessarily preferable to one with higher Precision, and vice versa. The appropriate balance between these metrics depends on the application and on the relative costs associated with failing to detect an expressed emotion, incorrectly identifying an emotion in a Neutral utterance, or confusing two distinct emotional categories.

\section{Conclusion and Future Work}
\label{conclusion}

We have previously presented a text classification approach to fill the gap between interpretability and performance, adapted for the task of ERC. This approach consisted of integrating Fuzzy Fingerprints (FFPs) into a PLM-based classifier, leveraging the advantages of both. We demonstrated that it is possible to achieve state-of-the-art results while also providing insights into the classification process, since our model’s decisions are grounded in measurable similarities to interpretable class prototypes. 

Results show that FFPs reduce the tendency to classify minority emotional utterances as \emph{Neutral}, a common issue in imbalanced datasets. This improvement, however, comes at the cost of a moderate increase in the classification of \emph{Neutral} utterances as emotional, reflecting the trade-off between Recall and Precision. 
	
The analysis of the confusion matrices further reveals that the nature of misclassifications differs across datasets, with \emph{Neutral} being the predominant destination of misclassified emotional utterances in DailyDialog, while confusion between distinct emotional categories is also more evident in MELD. 

Concerning the human evaluation study, annotators favored the outputs of our FFP model, suggesting that it captures emotional nuance in ways that resonate with human perception.

Our FFP mechanism is modular, making it suitable for integration with other types of classifiers, beyond PLM-based. We have done experiments concerning using FFPs directly on concatenated speech features for Cognitive Impairment Detection from speech \citep{botelho2025acoustic} in which the most difficult to detect minority class showed a great increase in F1 score.

For future work, we plan to test our approach with other types of classifiers and extend it beyond the task of ERC.

\printcredits
\section*{Acknowledgments}
\begin{sloppypar}
This work was supported by the Portuguese Recovery and Resilience Plan through project C645008882-00000055 (Responsible.AI); Fundação para a Ciência e a Tecnologia (FCT), through Portuguese national funds under projects UID/50021/2025 (DOI: \url{https://doi.org/10.54499/UID/50021/2025}) and UID/PRR/50021/2025 (DOI: \url{https://doi.org/10.54499/UID/PRR/50021/2025}) and grant UI/BD/154561/2022 (DOI: \url{https://doi.org/10.54499/UI/BD/154561/2022}); and partially by CLUL, UID/214/2025 (\url{https://doi.org/10.54499/UID/00214/2025}).
\end{sloppypar}
\section*{Declaration of generative AI and AI-assisted technologies in the manuscript preparation process}
During the preparation of this work the authors used ChatGPT in order to improve the readability of the manuscript. After using this tool, the authors reviewed and edited the content as needed and take full responsibility for the content of the published article.
\bibliographystyle{cas-model2-names}

\bibliography{cas-refs}



\end{document}